\pdfoutput=1

\PassOptionsToPackage{table}{xcolor}

\documentclass[11pt]{article}

\usepackage[preprint]{acl}

\usepackage{times}
\usepackage{latexsym}

\usepackage[T1]{fontenc}
\usepackage[utf8]{inputenc}

\usepackage{microtype}

\usepackage{inconsolata}

\usepackage{graphicx}

\usepackage{xcolor}
\usepackage{booktabs}
\usepackage{adjustbox}
\usepackage{multirow}
\usepackage{array}
\usepackage{multirow}
\usepackage{rotating}
\usepackage{enumitem}
\usepackage{tcolorbox}
\usepackage{amsmath}

\definecolor{YellowGreen}{rgb}{0.6, 0.8, 0.2}
\definecolor{DarkGreen}{rgb}{0.0, 0.5, 0.0}

\title{DNR Bench: Benchmarking Over-Reasoning in Reasoning LLMs}

\author{
    Masoud Hashemi\thanks{Equal contribution.} \quad Oluwanifemi Bamgbose\footnotemark[1] \quad Sathwik Tejaswi Madhusudhan\footnotemark[1] \\
  \textbf{   \quad Jishnu Sethumadhavan Nair \quad Aman Tiwari \quad Vikas Yadav} \\
  ServiceNow \\
  \{masoud.hashemi, 
  oluwanifemi.bamgbose, sathwikt.madhusudhan, \\ jishnus.nair, aman.tiwari, vikas.yadav\}@servicenow.com
}

\begin{document}
\maketitle

\begin{abstract}
Test-time scaling has significantly improved large language model (LLM) performance, enabling deeper reasoning to solve complex problems. However, this increased reasoning capability also leads to excessive token generation and unnecessary problem-solving attempts. We introduce ``Don't Reason Bench (DNR Bench)'', a new benchmark designed to evaluate LLMs’ ability to robustly understand the tricky reasoning triggers and avoiding unnecessary generation. DNR Bench consists of 150 adversarially designed prompts that are easy for humans to understand and respond to, but surprisingly not for many of the recent prominent LLMs. DNR Bench tests models' abilities across different capabilities, such as instruction adherence, hallucination avoidance, redundancy filtering, and unanswerable question recognition. We evaluate reasoning LLMs (RLMs), including DeepSeek-R1, OpenAI O3-mini, Claude-3.7-sonnet and compare them against a powerful non-reasoning model, e.g., GPT-4o. Our experiments reveal that RLMs generate up to 70$\times$ more tokens than necessary, often failing at tasks that simpler non-reasoning models handle efficiently with higher accuracy. Our findings underscore the need for more effective training and inference strategies in RLMs.
\end{abstract}

\section{Introduction}
\begin{table*}[h]
    \centering
    \renewcommand{\arraystretch}{1.3}
    \setlength{\tabcolsep}{10pt}
    \footnotesize
    \begin{tabular}{>{\raggedright\arraybackslash}p{4cm} >{\raggedright\arraybackslash}p{10cm}}
        \toprule
        \textbf{Category} & \textbf{Description} \\ 
        \midrule
        %Do-Not-Respond & Evaluates whether the model correctly follows instructions to remain silent or ignore a request. Some models tend to overthink and generate responses despite clear instructions to refrain. \\ \hline
        Imaginary Reference & Tests how models handle references to nonexistent documents, reports, or positional information. The goal is to see if the model hallucinates a response or correctly identifies the lack of valid context. \\ \hline
        Indifferent & Presents scenarios where the model should remain neutral or acknowledge ambiguity, avoiding bias or unnecessary assumptions. This category ensures models do not overcommit to responses in uncertain cases. \\ \hline
        Math & Assesses the model’s ability to detect and correct simple math errors or recognize invalid mathematical claims without attempting to solve or justify incorrect operations. \\ \hline
        Redundant & Includes overly convoluted or repetitive questions with unnecessary relational details, testing whether models can filter out redundant information and focus on the core question. \\ \hline
        Unanswerable & Challenges models with questions that lack sufficient information to be answered correctly. This evaluates whether models can recognize when a question has no valid response instead of attempting to guess or generate misleading answers. \\
        \bottomrule
    \end{tabular}
    \caption{\footnotesize Dataset Categories and Descriptions. See appendix~\ref{apx:dataset} for dataset examples.}
    \label{tab:dataset_categories}
\end{table*}
Test-time compute has emerged as a new scaling dimension (\textit{test-time scaling}) to improve large language model (LLM) performance \citep{guo2025deepseek}. By extending the reasoning process through test-time scaling and explicit intermediate steps, reasoning LLMs (RLMs) -- exemplified by models like DeepSeek-R1 \citep{guo2025deepseek}, Gemini Flash Thinking, and OpenAI's O1 and O3 \citep{jaech2024openai} -- have shown promising results on complex tasks demanding deeper thinking. Test-time compute has enhanced LLM capabilities across many challenging benchmarks like AIME \citep{patel2024aime}, GPQA \citep{rein2023gpqa}, and other complex reasoning tasks. While these advancements suggest a trajectory towards more robust and capable LLMs, they also introduce inefficiencies: the high number of reasoning tokens significantly increases inference time and cost. Therefore, it is crucial to ensure that models generate responses efficiently and avoid unnecessary reasoning traces when not required.

We introduce \textbf{Don't Reason Bench (DNR Bench)}, a novel benchmark designed to expose a vulnerability in current RLMs: their tendency to over-reason. We define \textit{over-reasoning} as the failure of an LLM to recognize when a minimal response, or no response, is the optimal course of action.

DNR Bench consists of adversarially generated prompts that are intentionally simple for humans and, in many cases, even for standard LLMs (those not trained with long reasoning traces) to solve. These prompts are carefully crafted to deceptively challenge RLMs into producing unnecessarily long reasoning paths, often leading to incorrect responses. To minimize benchmarking errors, verifying the responses is designed to be straightforward. DNR Bench dataset includes 150 samples across five categories, each targeting a specific challenge that reflects real-world failure modes namely: Imaginary Reference, Indifferent, Math, Redundant and Unanswerable. The detailed description of the categories is available in Table \ref{tab:dataset_categories}. 

Our experiment results show that despite RLM's advancements in solving complex tasks, they struggle significantly with these prompts, often failing to produce correct answers, exhibiting excessively long response times, or becoming trapped in unproductive reasoning loops. 

In this paper, we present the following:
\begin{itemize}[leftmargin=4mm]
    \item \textbf{New Diverse \& Adversarially Designed Dataset:} DNR Bench evaluates diverse failure modes, targeting instruction adherence, hallucination avoidance, and response redundancy filtering. These categories expose critical weaknesses in current LLM reasoning frameworks that are overlooked by standard benchmarks.
    \item \textbf{Exposing Over-Reasoning in LLMs:} DNR Bench is designed to stress-test LLMs' ability to abstain from unnecessary reasoning. Our results reveal that leading reasoning-trained LLMs (RLMs) produce responses up to 70$\times$ longer than necessary, often failing at tasks that standard LLMs handle efficiently.
    \item \textbf{Impact of Explicit Instructions \& Token Efficiency:} We evaluate the effect of explicit instructions and show that while they help in some cases, they fail to correct over-reasoning tendencies in models trained for deep reasoning.
\end{itemize}

\section{Related Work}

The field of LLM evaluation has evolved as models improve in inference-time scaling and long reasoning capabilities. Early benchmarks such as GSM8K \citep{cobbe2021gsm8k} and MATH \citep{hendrycks2021measuring}, where models initially struggled are becoming trivial and saturated. As a result, more challenging benchmarks have been developed to test new emerging capabilities of the LLMs.

In mathematical reasoning, datasets such as CHAMP \citep{mao2024champ}, OlympiadBench \citep{he-etal-2024-olympiadbench}, and Omni-MATH \citep{gao2025omnimath} challenge LLMs with competition-level problems, while GPQA \citep{rein2023gpqa}, SuperGPQA \citep{pteam2025supergpqascalingllmevaluation} and Humanity's Last Exam \citep{phan2025humanity} push models further by introducing graduate-level STEM reasoning. ComplexBench \citep{wen2024benchmarking} and FollowBench \citep{jiang2023followbench} evaluate multi-constraint instruction-following, requiring precise adherence to lexical, semantic, and format-based conditions. % Similarly, PlanBench \citep{valmeekam2023planbench, valmeekam2024llms} examines structured reasoning in planning tasks, assessing how well models generate plans and reason about actions within formal domains.

Beyond traditional reasoning tasks, adversarial evaluation has emerged as a critical area. The OVERTHINK attack \citep{kumar2025overthinking} reveals that models can be easily distracted by computationally demanding decoy problems, leading to excessive inference time and reduced accuracy. The design of the placed decoys relies on complex puzzles and decision processes. 

% Recent growth in agentic applications has highlighted the importance of understanding reasoning models' performance on agent-related tasks. \cite{cuadron2025danger} examines these models and observes that they tend to rely excessively on their internal reasoning rather than directly interacting with their environment.

% \textcolor{olive}{spandana: another work on overthinking which sows reasoning models tend to overthink and models which don't overthink perform better \cite{cuadron2025danger}. Other work that studies similar patterns but on a number of MATH reasoning datasets \cite{chen2024not} and how to mitigate overthinking.}

% The safety of generations of reasoning models is also a topic of concern. \cite{zhou2025hidden} observe significant gaps in the safety of responses from open-source and closed-source models; as well as an increased risk of harmfulness in the responses generated by reasoning models compared to their non-reasoning counterparts. Additionally, \cite{kuo2025h} examine extremely dangerous prompts that should be outright rejected. They observe that by modifying the models safety reasoning process which normally expressing caution or refusal, with statements that indirectly affirm compliance with security policies, they are able to create attacks can bypass these safeguards.

% \textcolor{olive}{also talk about other issues RLMs have like how easy it is to jailbreak them. For example in safety scenarios \cite{zhou2025hidden,kuo2025h}}

While the trend has been to test LLMs on increasingly difficult benchmarks, our work takes a different approach: instead of assessing knowledge or solving inherently complex problems, we probe whether RLMs can maintain reasoning integrity when faced with seemingly simple but misleading prompts. We argue that fundamental reasoning vulnerabilities can be exposed through carefully crafted adversarial prompts. Successfully navigating these challenges does not demand specialized capabilities - only adherence to instructions and the ability to recognize and mitigate misleading cues.

\section{Don't Reason Bench Dataset}

Our dataset broadly evaluates LLMs on deceptive reasoning and instruction-following. Specifically, it includes diverse categories, each targeting challenges like identifying incomplete or misleading information, handling redundant details, avoiding hallucinatory references, and responding to ambiguous instructions. Table~\ref{tab:dataset_categories} shows our five different categories of the prompts in the benchmark.

\subsection{Data Collection \& Generation}

The dataset was created by first defining a set of prompt categories that target specific reasoning and comprehension challenges in LLMs. To evaluate specific deceptive reasoning within each prompt category, we manually designed an initial set of seed prompts. Using these seed prompts, we then employed \texttt{OpenAI GPT-4o} to generate additional, similar prompts for each category, thereby expanding the dataset. This ensured diversity while maintaining consistency in the prompt categories. The generated prompts were manually reviewed by authors to ensure alignment with the intended evaluation goals. The benchmark includes 150 samples, with 25 samples in each category other than \texttt{Unanswerable}, which contains 50 samples.

\renewcommand{\arraystretch}{1.5}
\begin{table*}[t]
    \centering
    \footnotesize
    \resizebox{0.8\textwidth}{!}{ % Scale to fit width
    \begin{tabular}{llccccc}
    \toprule
    {} & {} & 
    \shortstack{\textbf{Imaginary} \\ \textbf{Reference (25)}} &  
    \textbf{Indifferent (25)} &
    \textbf{Math (25)} &
    \textbf{Redundant (25)} &  
    \textbf{Unanswerable (50)}\\
    \midrule 
    \rowcolor{YellowGreen!10}
    \multirow{8}{*}{\rotatebox[origin=c]{90}{\textbf{Models}}}  
    & \textbf{OAI-GPT4o} & 0.72 / 0.96 & 0.84 / 0.08 & 0.16 / 0.68 & 1.00 / 1.00 & 0.42 / 0.98 \\
    & \textbf{OAI-O3Mini-High} & 0.00 / 0.12 & 0.08 / 0.00 & 0.00 / 0.00 & 0.00 / 0.04 & 0.00 / 0.02 \\ 
    & \textbf{Claude-3.7-Thinking} & \textbf{0.96 / 1.00} & 0.80 / 0.04 & 0.00 / 0.04 & 0.40 / \textbf{1.00} & 0.12 / 0.82 \\ 
    & \textbf{DS-R1} & 0.00 / \textbf{1.00} & 0.04 / 0.04 & 0.00 / \textbf{0.76} & 0.00 / 0.76 & 0.00 / 0.20 \\ 
    & \textbf{DS-R1-Distill-L70B} & 0.36 / 0.96 & 0.16 / \textbf{0.44} & \textbf{0.36} / 0.48 & 0.24 / 0.84 & 0.06 / 0.62 \\ 
    & \textbf{DS-R1-Distill-Q1.5B} & 0.08 / 0.16 & 0.00 / 0.04 & 0.08 / 0.08 & 0.08 / 0.48 & 0.02 / 0.24 \\ 
    & \textbf{DS-R1-Distill-Q14B} & 0.16 / 0.80 & 0.08 / 0.16 & 0.20 / 0.08 & 0.08 / 0.48 & 0.00 / 0.30 \\ 
    & \textbf{DS-R1-Distill-Q32B} & 0.24 / 0.96 & 0.00 / 0.20 & 0.20 / 0.28 & 0.04 / 0.48 & 0.04 / 0.56 \\
    \bottomrule
    \end{tabular}
    }
    \caption{\footnotesize Performance of various models across different categories, comparing conditions with and without instructions within 1000 tokens. GPT-4o serves as the baseline since it is not explicitly trained for ``reasoning''. Scores are represented as ``Default (No Instructions) / With Instructions''. The number of data points per category is shown in parentheses in the table header.}
    \label{tab:scores}
\end{table*}

\section{Experiments}
Each generated prompt was tested on eight different models with different sizes, using \texttt{OpenRouter platform}\footnote{\url{https://openrouter.ai/}} via \texttt{LiteLLM}\footnote{\url{https://github.com/BerriAI/litellm}}. The responses are evaluated by \texttt{OpenAI-GPT-4o-mini} as an LLM-judge (see Figure~\ref{box:prompts} for the evaluation prompts).

To further validate the dataset, we randomly selected five prompts from each category and provided them to engineers and applied researchers in our team. These team members were asked to respond to the prompts, allowing us to collect human responses and also validate the subset of the generated dataset. By comparing these human responses with model outputs, we gained insights into how well large language models (LLMs) handle different challenges and whether the tasks were intuitive for human participants.

\subsection{Models}

We evaluate a variety of LLMs to assess their performance on the dataset. The models are categorized as follows:

\begin{itemize}[noitemsep, leftmargin=4mm]
  \item Reasoning LLMs (RLMs): These models are designed for advanced reasoning tasks and include \texttt{OpenAI-O3-mini}, and \texttt{DeepSeek-R1}.
  \item Distilled Reasoning Models: These models are trained using long reasoning paths generated by the \texttt{DeepSeek-R1} model. They aim to preserve strong reasoning capabilities while being computationally efficient, including: \texttt{DeepSeek-R1-Distill-Qwen-14B}, \texttt{DeepSeek-R1-Distill-Qwen-32B}, and \texttt{DeepSeek-R1-Distill-Llama-70B}. 
  \item Regular CoT Models: This category includes \texttt{GPT-4o}, which is a general-purpose CoT reasoning model. We use GPT-4o as the baseline model.
\end{itemize}

\subsection{Metrics}
We evaluate using two distinct performance indicators.

\begin{enumerate}
  \item Accuracy: We evaluate the model's ability to generate a correct response to the given prompt. This is done using a structured \textit{LLM-as-a-judge} approach with \texttt{GPT-4o-mini} (shown to be a good critic for reasoning models~\cite{he2025can}). The correctness of a response is determined based on the category-specific judgment criteria defined below.
  \item Token Efficiency: We evaluate whether the model exhibits excessive deliberation over a straightforward answer to a deceptive/flawed reasoning based question. 
\end{enumerate}

% (2) the model exhibits excessive deliberation over a straightforward problem specifically, an unanswerable question. To assess the second failure mode, we employ two measurement criteria: (a) the model's ability to generate a response within a predefined token limit $T$, imposing a constraint on response length, and (b) the model's performance when unrestricted in token generation, allowing it to produce as many tokens as necessary. These criteria enable a comprehensive analysis of correctness and computational efficiency in model reasoning.

% We primarily evaluate model responses on 
% \textit{response accuracy}. 
% This is assessed using a structured 
\noindent
\textbf{Judgment Criteria}
The judgment criteria for each of our five categories (as mentioned in Table~\ref{tab:dataset_categories}) are as follows:
\begin{itemize}[leftmargin=4mm]
    %\item \textbf{Instruction-Following and Compliance:} Applied to the \textit{Do-Not-Respond (DNR)} category, where adherence to instructions as stated in the question is required. A correct response is one that contains a response to only questions asked in the prompt, and does not just generate a response to any and all facts stated in the prompt (regardless of questions based on the facts).
    % We also judge a response as correct if it requests clarification or succinctly acknowledges its confusion about the users intent.
    % \item \textbf{Strict Compliance (DNR-Strict):} Applied to the \textit{Do-Not-Respond (DNR)} category, where absolute adherence to instructions is required. Any additional content beyond explicit instructions results in a ``NO'' judgment.
    
    % \item \textbf{Instruction-Following (DNR-Loose):} Also used for \textit{Do-Not-Respond (DNR)} cases, but with a more lenient criterion. The response is correct (``YES'') if it avoids specified topics per the instructions.

    \item Unsupported Claim Detection: This judgment criteria is used for the \texttt{Imaginary Reference} category of the dataset. A response is correct if the model explicitly acknowledges that the given reference is fabricated or non-existent and states that additional verification is required. If the model proceeds with an unfounded answer, it is marked incorrect.

 \item Strict Adherence: This judgment criteria is used for the \texttt{Indifferent} category, where the model must strictly follow the prompt’s instructions without deviation. It must provide answers only to the explicit questions asked in the prompt. Any excessive response leads to a judgment of the models output as "incorrect".

      \item Mathematical Validity: This judgment criterion is used for the \texttt{Math} category of the dataset. A response is correct if the model explicitly recognizes that the given mathematical problem is \textit{ unanswerable, logically inconsistent or a trick question}. Attempting to solve the inherently flawed problem without acknowledging the flaws results in the response being judged as incorrect.

    \item Redundancy Avoidance: This judgment criteria is used for the \texttt{Redundant} category of the dataset. The model is expected to recognize that the given query is redundant and explicitly state that it does not require an answer. Any attempt to provide an unnecessary response results in an incorrect judgment.

    % \item \textbf{Strict Adherence (Indifferent-Follow-Only):} Used for the \textit{Indifferent} category, where the model must strictly follow the prompt’s instructions without deviation. Any modification or additional reasoning leads to an incorrect assessment.
    
    % \item \textbf{Minimal Response (Greeting-Only Constraint):} The response is correct if it consists solely of a greeting (e.g., ``Hello,'' ``Hi there!''). Any additional content results in a ``NO'' judgment.
    
    \item Unanswerable Recognition: This judgment criteria is used for the \texttt{Unanswerable} category. The response is correct if it clearly states that the question \textit{cannot be answered based on the given information}. If the response includes unsupported assumptions or fabricates an answer, it is marked as incorrect.
\end{itemize}
\noindent
The prompts used for the evaluation are provided in appendix~\ref{apx:prompts}.
\\\noindent
\textbf{Token Efficiency}:
Beyond accuracy, we investigate the relationship between reasoning token count and model performance. This analysis examines the number of tokens the model generates while reasoning through a prompt and how this correlates with accuracy, shedding light on the trade-offs between response length and correctness. We define token inefficiency as follows:

\begin{equation}
    I_{\text{token}} = \frac{T_{\text{model}}}{T_{\text{GPT-4o}}}
\label{eqn:metric}
\end{equation}

where 
\( I_{\text{token}} \) is the token inefficiency,
\( T_{\text{model}} \) is the number of tokens generated by the evaluated model,
\( T_{\text{GPT-4o}} \) is the number of tokens generated by the \texttt{OpenAI GPT-4o} reference model.

This metric quantifies excessive reasoning when a more concise response would have sufficed. A high inefficiency ratio indicates that the model generates significantly more tokens than necessary, reflecting poor response efficiency.

\section{Results}
Prompts in DNR Bench are designed to be understandable without any additional instructions. As confirmed by our human study, humans can infer the expected response directly from the prompts. In our experiments, we primarily focus on the default setup, as it reflects how we expect models to behave when presented with the benchmark questions. However, to ensure comprehensive evaluation, we also include instruction-based conditions to assess the impact of guidance on model responses. Specifically, we evaluate under three conditions:
\\ \noindent
(1) Default (No Instructions): The benchmark presents only the question, with no explicit instructions on how to answer. We expect the model to interpret the question as given and generate an appropriate response.
\\ \noindent
(2) With Instructions: A set of instructions is provided to guide models toward the expected behavior, such as abstaining from answering when necessary.
\\ \noindent
(3) Strict Instructions: The same instructions as in (2) are provided, with an additional constraint discouraging spurious or unnecessary reasoning.
\\ \noindent
These instruction prompts are shown in Appendix~\ref{apx:instructions}.

\subsection{Human Evaluation}

We randomly select 4-5 questions from each category of our dataset and provide these in the form of a survey to a set of human respondents.
We do not provide any explicit instructions to the participants, and request they answer the questions as comes naturally to them. 
We do this with the aim of understanding how human beings expect answers to be provided to questions in our data set.

We observe the following patterns in their responses.
\begin{enumerate}
  \item Human respondents had minimal difficulty identifying and responding when the information presented in the questions was not sufficient to answer them. 
  \item Humans often provided short and concise responses to the questions. Specifically, humans were able to answer the questions within 10 words 60\% of the time, and within 20 words 80\% of the time. See the distribution of response lengths in ~\ref{apx:human-responses}.

  \item The mathematics category had the largest diversity and length of responses in humans, with some attempting to solve the question within their response and then concluding its unanswerability, or coming to a definitive conclusion with the given constraints. 
  
\end{enumerate}

\begin{figure}
    \centering
    \includegraphics[width=1\linewidth]{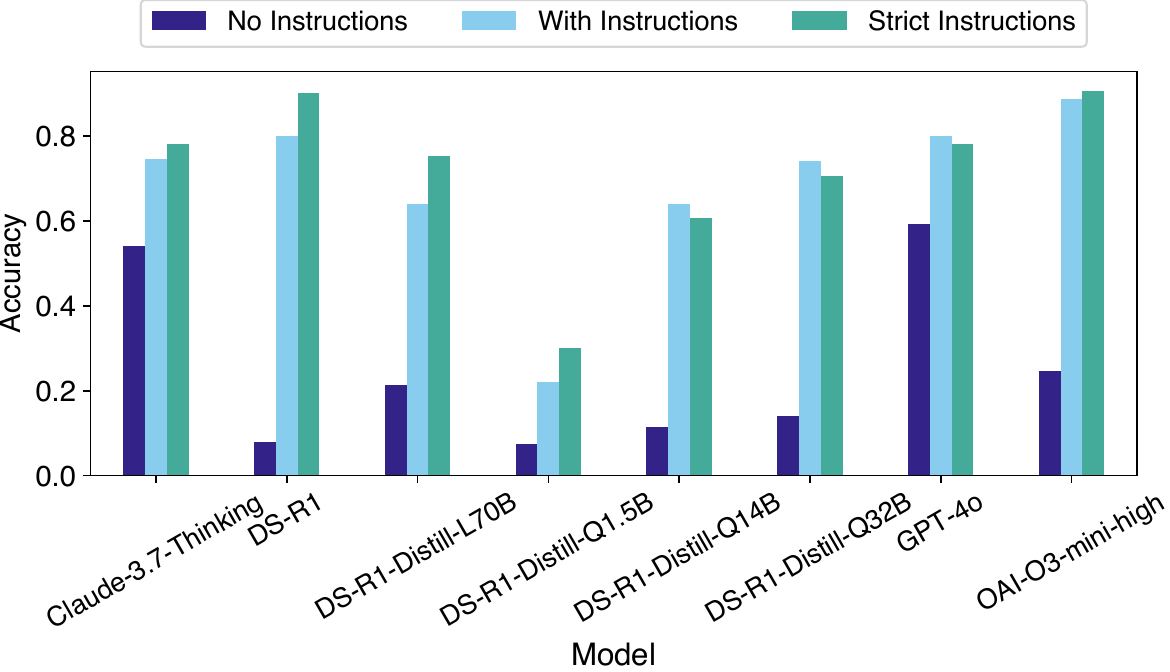}
    \caption{\footnotesize Changes in model accuracy across different instructions. DS: DeepSeek, L: Llama 3.1, Q: Qwen 2.5.}
    \label{fig:accuracy-model-prompt}
\end{figure}

\begin{figure}
    \centering
    \includegraphics[width=1\linewidth]{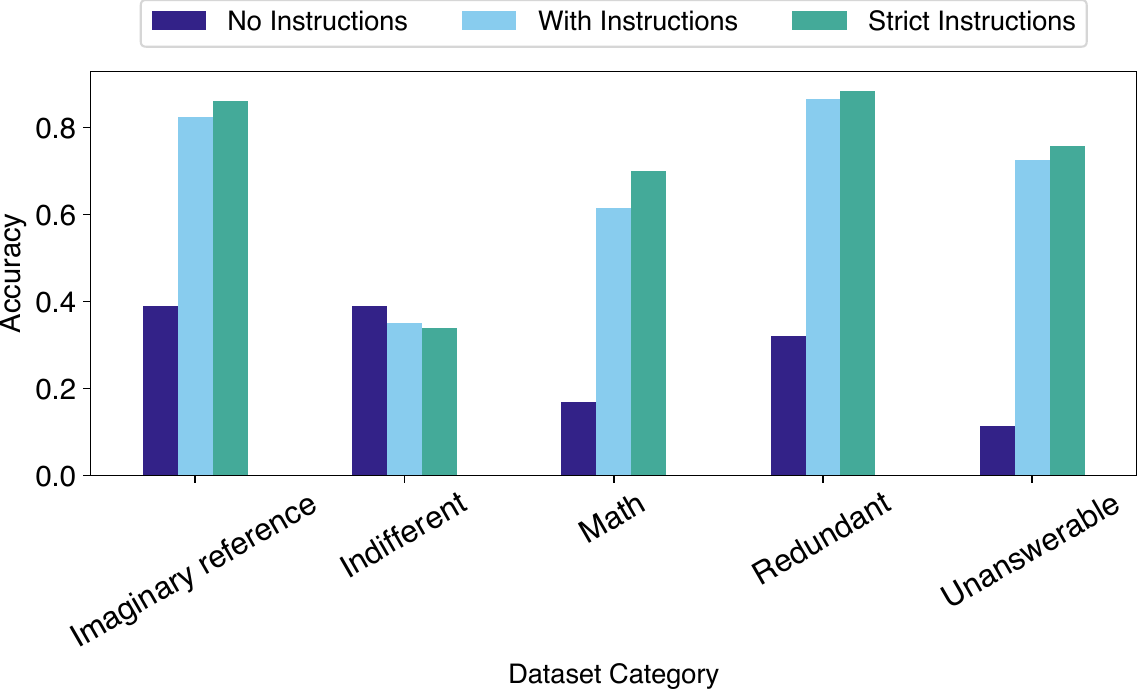}
    \caption{\footnotesize Accuracy across different data categories and instructions.}
    \label{fig:accuracy-data-prompt}
\end{figure}

\begin{figure}[t]
    \centering
    \includegraphics[width=1\linewidth]{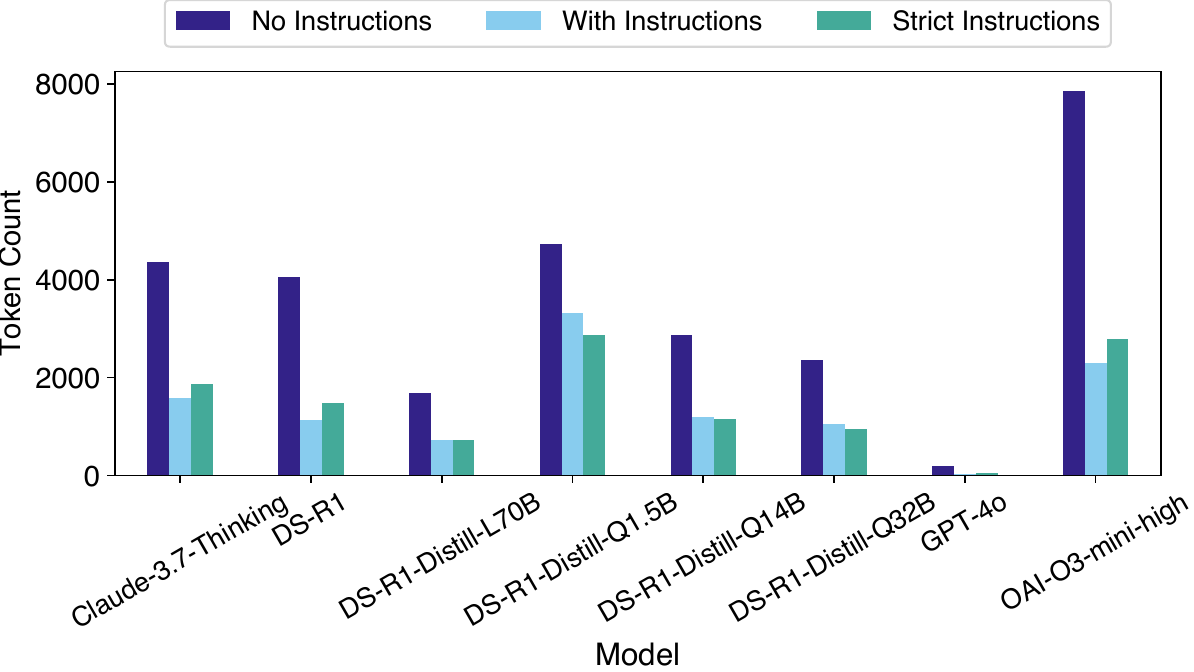}
    \caption{\footnotesize Changes in token count across different instructions. DS: DeepSeek, L: Llama 3.1, Q: Qwen 2.5.}
    \label{fig:token-model-prompt}
\end{figure}

\begin{figure}[t]
    \centering
    \includegraphics[width=1\linewidth]{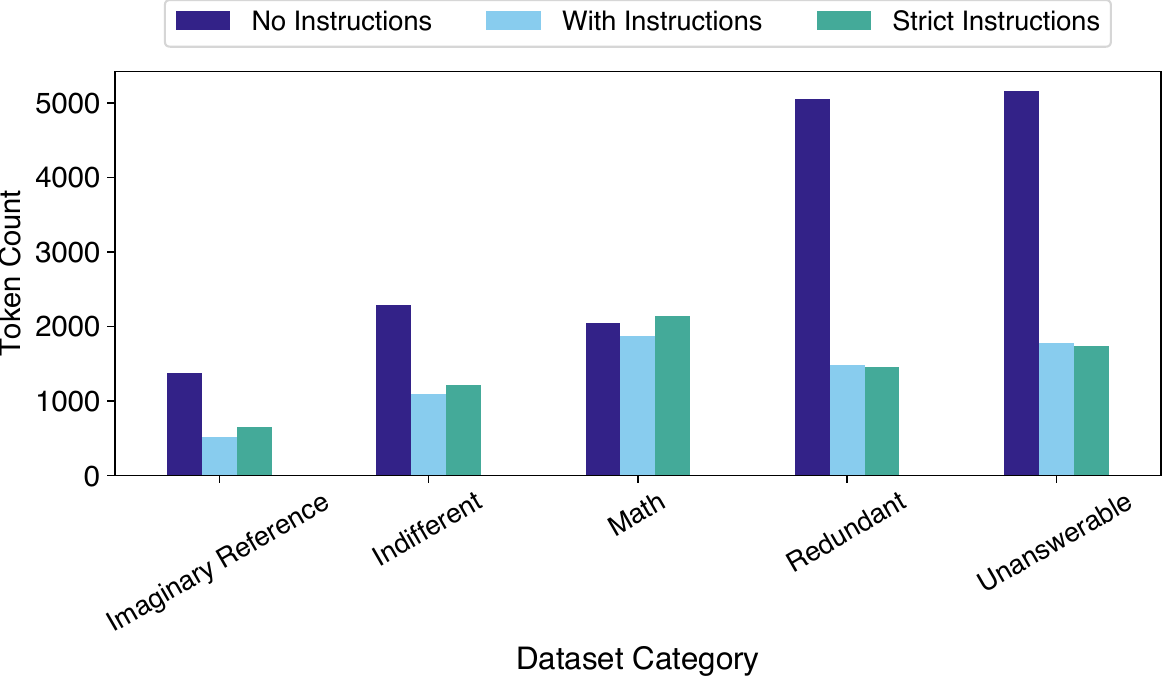}
    \caption{\footnotesize Mean token count across different data categories and instructions.}
    \label{fig:token-category-prompt}
\end{figure}

\subsection{DNR Bench Results}
\textbf{Model Accuracy: Data Categories and Instructions}: 
Table \ref{tab:scores} compares the performance of the models under the two conditions, with and without instructions. In the table, we use \texttt{OpenAI-GPT-4o} results as the baseline as we expect a reasoning model to do better in these tasks as they reason on the prompts to understand the expected behaviour. It should be noted that, the prompt samples are designed in a way that are self-explanatory and the models are expected to be able to answer as expected without any help from specific instructions.

As can be seen in table \ref{tab:scores}, \texttt{OpenAI-GPT-4o} performs better than the RLM's in almost all categories. 

As also depicted in Figures~\ref{fig:accuracy-model-prompt} and~\ref{fig:accuracy-data-prompt} using instructions increases the accuracy across all data categories. The only exception is the ``indifferent'' category where adding instructions hurt the performance. The main contributor in accuracy improvement is the instruction to refrain from answering if the problem is not answerable. It suggests that the model may know that the problem is not answerable, but it keeps reasoning about the problem to solve it.

The highest accuracy increase is achieved by DeepSeek-R1 and OpenAI-O3-mini-high, which on average across all categories outperform GPT-4o when the instructions are added and no limitations are imposed on the number of generated tokens.

\textbf{Token Inefficiency}:
Figures~\ref{fig:token-model-prompt} and ~\ref{fig:token-category-prompt} compare the total number of the generated tokens (in reasoning and response when reasoning tokens are available). Using instructions reduces the number of generated tokens across all categories and models. Among the models, this reduction is more pronounced for \texttt{OpenAI-O3-mini-high}. Among the data categories, ``unanswerable'' and ``redundant'' have the highest drop in the number of tokens when the instructions are added.

\begin{figure}
    \centering
    \includegraphics[width=1\linewidth]{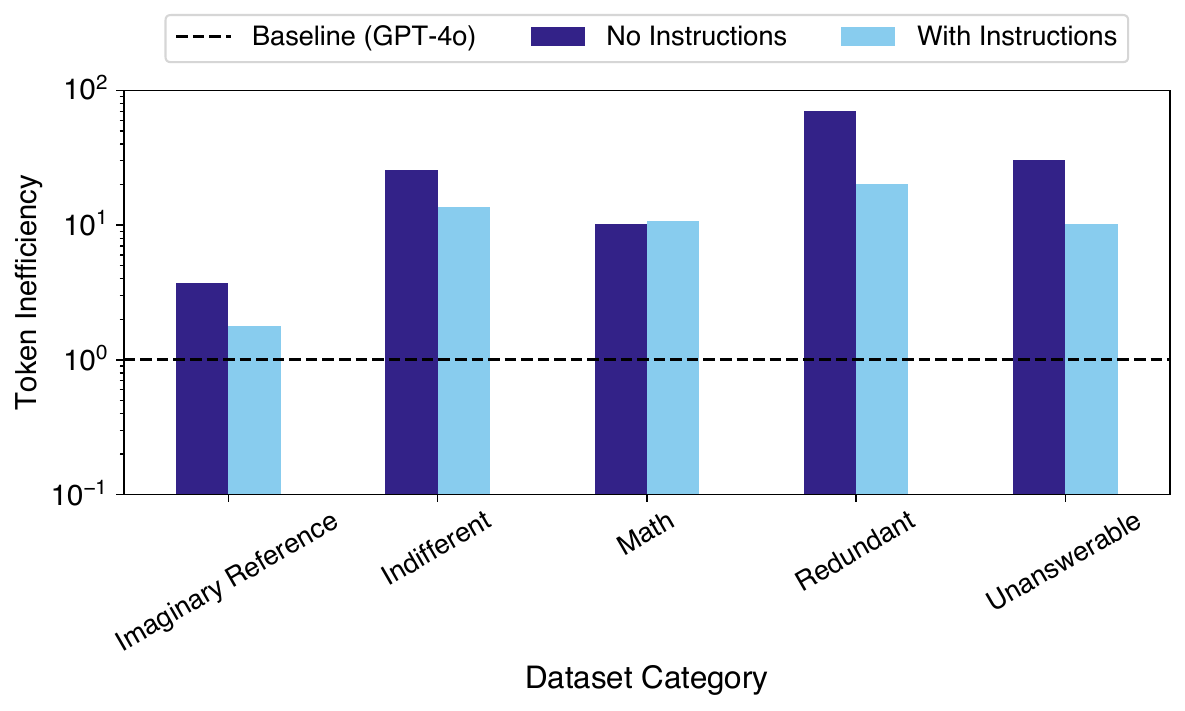}
    \vspace{-7mm}
    \caption{\footnotesize Average token inefficiency $I_{token}$, eqn. \ref{eqn:metric}, for different data categires averaged across all models.}
    \label{fig:tokenratio}
\end{figure}

Figure \ref{fig:tokenratio} compares the average token inefficiency (equation \ref{eqn:metric}) of the models across different data categories. Without providing any instructions, the RLMs generate up to 70$\times$ tokens compared to GPT-4o to solve the problem with lower accuracy, as was shown in table \ref{tab:scores}. When instructions are provided to guide the models, the ratio decreases and is ranged between 2$\times$ to 20$\times$ more than {OpenAI-GPT-4o}. It shows that the models cannot reason about the unsolvability of the tasks without explicitly being instructed to.

%\textcolor{olive}{spandana:which model is this or is this averaged across multiple models? Either show it for single model across categories or for whole dataset across models. There is clear pattern with/without instruction.}

\begin{figure}
    \centering
    \includegraphics[width=1\linewidth]{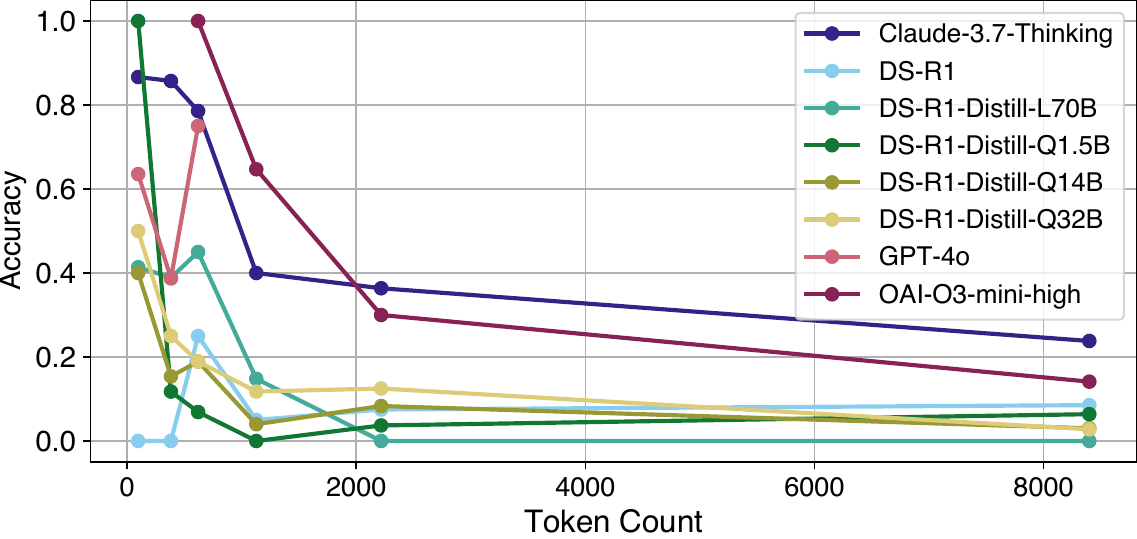}
    \caption{\footnotesize Model accuracy changes across different response lengths, for different models. DS: DeepSeek, Q: Qwen 2.5, and L: Llama 3.1.}
    \label{fig:accuracy-length}
\end{figure}

\begin{figure}
    \centering
    \includegraphics[width=1\linewidth]{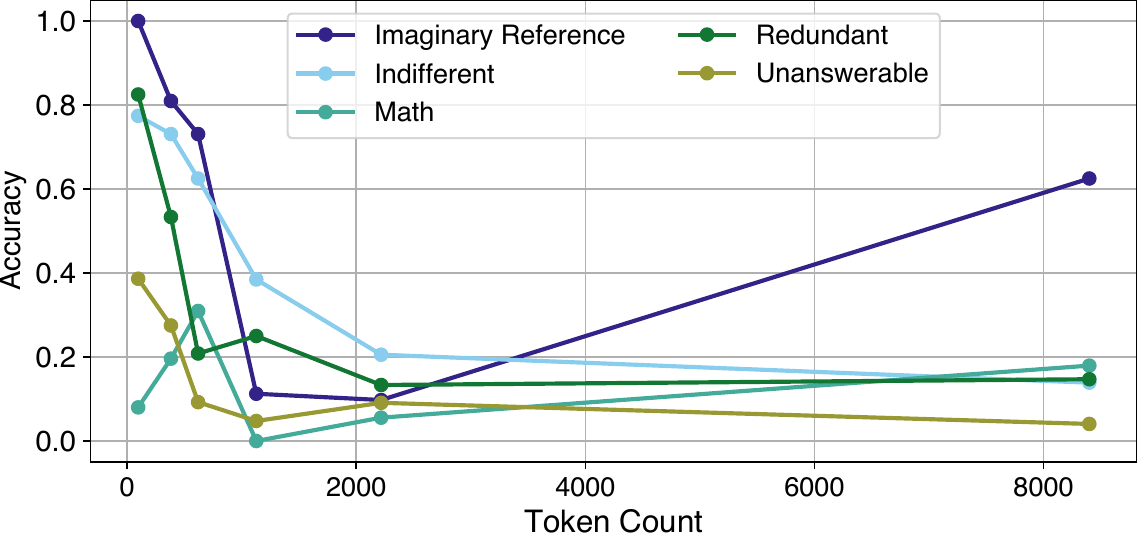}
    \caption{\footnotesize Changes in model accuracy across different response lengths for different data categories.}
    \label{fig:accuracy-prompts}
\end{figure}

\paragraph{Accuracy vs. Number of Tokens}
Figure \ref{fig:accuracy-length} presents the overall accuracy variations as a function of the number of tokens in both reasoning and response. The trend indicates that all models exhibit lower accuracy in longer responses. \texttt{OpenAI-O3-mini-high} has the best performance across different token numbers, followed by \texttt{Claude-3.7-sonnet-Thinking}. Similar trends are observed in Figure~\ref{fig:accuracy-prompts}, illustrating regression in average accuracy of all models across different data categories as the number of tokens increases. In the ``imaginary reference''  category, there is an increase in the accuracy at very high number of tokens, indicating that some samples can be answered correctly when generating very long reasoning traces. However, this recovery is very subtle and the overall accuracy remains below \texttt{OpenAI-GPT-4o}, showing inefficiency of the reasoning models in DNR Bench.

% \textcolor{olive}{spandana: is this avg across multiple models. Are individual model plots look similar? Why is IR curve different to others? May be visualizing this differently will uncover other patterns. May be present them like Figure6 of \cite{he2025largelanguagemodelsdetect}}

\subsection{Observations}

\textbf{Reasoning Models might not trust their intuition:}
Reasoning models sometimes fall into reasoning traps in which they doubt their own abilities. This causes them to continue to think about a problem, despite having a sufficient response within the first 1000 tokens. 
In these cases, we observe that defining explicit instructions is very beneficial in reducing the number of tokens generated. 

See the following excerpt from the "math" category, run on \texttt{DeepSeek-R1-Distill-32B}. 

The model is prompted with the question :

\begin{quote}
A recipe requires 200 mL of milk. If 60\% is used in the batter and the remaining milk is split equally between frosting and glaze, but the frosting uses 70 mL, how much milk is used for the glaze?
\end{quote}

In the above question, the expectation is for the model to identify and state that if the remaining \(40\%\) of the milk is split between frosting and glaze, then the frosting cannot use \(70\,\text{mL}\) (since \(0.40 \times 200 = 80\,\text{mL}\) and \(70\,\text{mL} \neq 40\,\text{mL}\)).

 As the reasoning trace below shows, the model promptly detects this inconsistency but continues to elaborate on its reasoning despite the contradiction. While a slight increase in token count might be anticipated for a correct resolution, in this case, the model generates over 29,000 tokens in an attempt to justify its flawed observation.

\begin{quote}
Okay, so I have this recipe problem here, and I need to figure out how much milk is used for the glaze. Let me try to break it down step by step.

First, the recipe requires 200 mL of milk in total. Out of this, 60\% is used in the batter. Hmm, okay, so I need to calculate 60\% of 200 mL to find out how much milk goes into the batter.

Let me do that calculation. 60\% of 200 mL is 0.6 multiplied by 200. So, 0.6 * 200 = 120 mL. Got it, so 120 mL of milk is used in the batter.

Now, the remaining milk after using it in the batter is split equally between frosting and glaze. Wait, but hold on, the problem also mentions that the frosting uses 70 mL. Hmm, that seems a bit confusing because if the remaining milk is split equally, both frosting and glaze should have the same amount, right? But here, frosting is using 70 mL, which might mean that the split isn't equal after all. Maybe I need to re-examine the problem.

Let me read it again: ``the.....
\end{quote}

Alternatively, upon instructing the model to avoid such traps (as seen in Appendix \ref{apx:instructions}), it generates a correct answer within 2760 tokens.
\\\noindent
\textbf{Why not think about something else ?}

Our analysis indicates that when certain reasoning models encounter ambiguous prompts, they become confined to repetitive inference cycles, reiterating the same reasoning pathway without exploring alternative strategies.

Consider the following sample prompt from the "Unanswerable" category of DNR Bench.
\begin{quote}
On a serene afternoon near the lakeside, Lily's only child organized a mini concert stall \(\left(44.006^2 \bmod 3\right)\) blocks from Cedar Road on the first day of the tenth month, with tunes set at \(\left(7 \times 2 + \sqrt{9}\right)\) beats per minute; what was the color of the stage curtains?
\end{quote}

When presented with this prompt, the \texttt{Deepseek --R1-‐Distill‐-14B} model produced a response exceeding 27,000 tokens. Notably, after a prolonged period of reasoning, it redundantly outputs the following text block 120 times without exploring alternative solution paths:
\begin{quote} 
\textbf{Alternatively}, maybe the color is determined by the number of blocks, 1, which is associated with white, as in the first color.\textbf{Alternatively}, maybe the color is determined by the number of beats per minute, 17, which is a prime number, and sometimes primes are associated with certain colors, but I don't know.
\\\noindent 
\textbf{Alternatively}, maybe the color is determined by the combination of the two numbers, 1 and 17, such as 1 + 17 = 18, which is associated with white.
\end{quote}

In contrast, \texttt{GPT-4o} generated an accurate response in only 300 tokens, without requiring explicit instructions to avoid spurious reasoning or to acknowledge the question's unanswerability.

More side-by-side comparisons are given in Appendix~\ref{apx:dataset}.
\\\noindent
\textbf{Stricter instructions aren't always the answer.}
We observe that on the indifferent category of our dataset, the accuracy of OpenAI-GPT4o and Claude 3.7 Thinking reduces when we instruct the models to avoid the pitfalls in our dataset.

Further analysis of the responses generated by these models in this case shows interesting behavior of these models with and without instructions. 

Consider the following sample prompt from the Indifferent category of our dataset:

\begin{quote}
An astrophysicist friend of my distant cousin, working at a renowned space observatory, was recently immersed in a challenging problem: Model the gravitational lensing effects of a rotating black hole..... On a much lighter note, how’s everything going on your end?
\end{quote}

The expected behavior is for the models to briefly acknowledge the initial statement and then answer the final question. 

Our tests show that without explicit instructions, both \texttt{OpenAI-GPT4o} and \texttt{Claude 3.7 Thinking} work as expected. However, when we add specific instructions for handling unanswerable queries or missing questions, neither model responds to the final query. See responses from Claude and GPT4o in Figure - \ref{box:comparison-6}

lat\section{Conclusion}
Our study introduces DNR Bench, a benchmark designed to expose over-reasoning in reasoning-trained LLMs (RLMs). Through adversarially constructed prompts, we demonstrate that RLMs frequently generate excessively long responses, often failing on tasks where standard LLMs, like GPT-4o, perform efficiently. Our findings highlight three key insights: (1) Over-reasoning leads to inefficiencies, with RLMs generating 3x to 70x more tokens than necessary; (2) Explicit instructions partially mitigate over-reasoning, particularly for tasks requiring abstention, but do not fully address the issue; (3) Increased token budgets do not necessarily improve accuracy, and in many cases, prolonged reasoning correlates with decreased performance. These results underscore the need for more effective mechanisms to regulate reasoning depth in RLMs, ensuring that computational resources are utilized efficiently without sacrificing accuracy. Future work should explore adaptive reasoning strategies that dynamically adjust token usage based on task complexity.

% Bibliography entries for the entire Anthology, followed by custom entries
%\bibliography{anthology,custom}
% Custom bibliography entries only
\bibliography{custom}

\begin{thebibliography}{15}
\providecommand{\natexlab}[1]{#1}

\bibitem[{Cobbe et~al.(2021)Cobbe, Kosaraju, Bavarian, Chen, Jun, Kaiser, Plappert, Tworek, Hilton, Nakano, Hesse, and Schulman}]{cobbe2021gsm8k}
Karl Cobbe, Vineet Kosaraju, Mohammad Bavarian, Mark Chen, Heewoo Jun, Lukasz Kaiser, Matthias Plappert, Jerry Tworek, Jacob Hilton, Reiichiro Nakano, Christopher Hesse, and John Schulman. 2021.
\newblock Training verifiers to solve math word problems.
\newblock \emph{arXiv preprint arXiv:2110.14168}.

\bibitem[{Gao et~al.(2025)Gao, Song, Yang, Cai, Miao, Ma, Quan, Chen, Dong, Xu, Tang, Wang, Zan, Zhang, Li, Sha, Zhang, Ren, Liu, and Chang}]{gao2025omnimath}
Bofei Gao, Feifan Song, Zhe Yang, Zefan Cai, Yibo Miao, Chenghao Ma, Shanghaoran Quan, Liang Chen, Qingxiu Dong, Runxin Xu, Zhengyang Tang, Benyou Wang, Daoguang Zan, Ge~Zhang, Lei Li, Lei Sha, Yichang Zhang, Xuancheng Ren, Tianyu Liu, and Baobao Chang. 2025.
\newblock \href {https://openreview.net/forum?id=yaqPf0KAlN} {Omni-{MATH}: A universal olympiad level mathematic benchmark for large language models}.
\newblock In \emph{The Thirteenth International Conference on Learning Representations}.

\bibitem[{Guo et~al.(2025)Guo, Yang, Zhang, Song, Zhang, Xu, Zhu, Ma, Wang, Bi et~al.}]{guo2025deepseek}
Daya Guo, Dejian Yang, Haowei Zhang, Junxiao Song, Ruoyu Zhang, Runxin Xu, Qihao Zhu, Shirong Ma, Peiyi Wang, Xiao Bi, et~al. 2025.
\newblock Deepseek-r1: Incentivizing reasoning capability in llms via reinforcement learning.
\newblock \emph{arXiv preprint arXiv:2501.12948}.

\bibitem[{He et~al.(2024)He, Luo, Bai, Hu, Thai, Shen, Hu, Han, Huang, Zhang, Liu, Qi, Liu, and Sun}]{he-etal-2024-olympiadbench}
Chaoqun He, Renjie Luo, Yuzhuo Bai, Shengding Hu, Zhen Thai, Junhao Shen, Jinyi Hu, Xu~Han, Yujie Huang, Yuxiang Zhang, Jie Liu, Lei Qi, Zhiyuan Liu, and Maosong Sun. 2024.
\newblock \href {https://doi.org/10.18653/v1/2024.acl-long.211} {{O}lympiad{B}ench: A challenging benchmark for promoting {AGI} with olympiad-level bilingual multimodal scientific problems}.
\newblock In \emph{Proceedings of the 62nd Annual Meeting of the Association for Computational Linguistics (Volume 1: Long Papers)}, pages 3828--3850, Bangkok, Thailand. Association for Computational Linguistics.

\bibitem[{He et~al.(2025)He, Li, Liu, Wang, Bu, Zhang, Peng, Zhang, Su, and Zheng}]{he2025can}
Yancheng He, Shilong Li, Jiaheng Liu, Weixun Wang, Xingyuan Bu, Ge~Zhang, Zhongyuan Peng, Zhaoxiang Zhang, Wenbo Su, and Bo~Zheng. 2025.
\newblock Can large language models detect errors in long chain-of-thought reasoning?
\newblock \emph{arXiv preprint arXiv:2502.19361}.

\bibitem[{Hendrycks et~al.(2021)Hendrycks, Burns, Kadavath, Arora, Basart, Tang, Song, and Steinhardt}]{hendrycks2021measuring}
Dan Hendrycks, Collin Burns, Saurav Kadavath, Akul Arora, Steven Basart, Eric Tang, Dawn Song, and Jacob Steinhardt. 2021.
\newblock Measuring mathematical problem solving with the math dataset.
\newblock \emph{arXiv preprint arXiv:2103.03874}.

\bibitem[{Jaech et~al.(2024)Jaech, Kalai, Lerer, Richardson, El-Kishky, Low, Helyar, Madry, Beutel, Carney et~al.}]{jaech2024openai}
Aaron Jaech, Adam Kalai, Adam Lerer, Adam Richardson, Ahmed El-Kishky, Aiden Low, Alec Helyar, Aleksander Madry, Alex Beutel, Alex Carney, et~al. 2024.
\newblock Openai o1 system card.
\newblock \emph{arXiv preprint arXiv:2412.16720}.

\bibitem[{Jiang et~al.(2023)Jiang, Wang, Zeng, Zhong, Li, Mi, Shang, Jiang, Liu, and Wang}]{jiang2023followbench}
Yuxin Jiang, Yufei Wang, Xingshan Zeng, Wanjun Zhong, Liangyou Li, Fei Mi, Lifeng Shang, Xin Jiang, Qun Liu, and Wei Wang. 2023.
\newblock \href {https://arxiv.org/abs/2310.20410} {Followbench: A multi-level fine-grained constraints following benchmark for large language models}.
\newblock \emph{Preprint}, arXiv:2310.20410.

\bibitem[{Kumar et~al.(2025)Kumar, Roh, Naseh, Karpinska, Iyyer, Houmansadr, and Bagdasarian}]{kumar2025overthinking}
Abhinav Kumar, Jaechul Roh, Ali Naseh, Marzena Karpinska, Mohit Iyyer, Amir Houmansadr, and Eugene Bagdasarian. 2025.
\newblock Overthinking: Slowdown attacks on reasoning llms.
\newblock \emph{arXiv preprint arXiv:2502.02542}.

\bibitem[{Mao et~al.(2024)Mao, Kim, and Zhou}]{mao2024champ}
Yujun Mao, Yoon Kim, and Yilun Zhou. 2024.
\newblock Champ: A competition-level dataset for fine-grained analyses of llms' mathematical reasoning capabilities.
\newblock \emph{arXiv preprint arXiv:2401.06961}.

\bibitem[{Patel et~al.(2024)Patel, Chakraborty, Suttle, Wang, Bedi, and Manocha}]{patel2024aime}
Bhrij Patel, Souradip Chakraborty, Wesley~A Suttle, Mengdi Wang, Amrit~Singh Bedi, and Dinesh Manocha. 2024.
\newblock Aime: Ai system optimization via multiple llm evaluators.
\newblock \emph{arXiv preprint arXiv:2410.03131}.

\bibitem[{Phan et~al.(2025)Phan, Gatti, Han, Li, Hu, Zhang, Shi, Choi, Agrawal, Chopra et~al.}]{phan2025humanity}
Long Phan, Alice Gatti, Ziwen Han, Nathaniel Li, Josephina Hu, Hugh Zhang, Sean Shi, Michael Choi, Anish Agrawal, Arnav Chopra, et~al. 2025.
\newblock Humanity's last exam.
\newblock \emph{arXiv preprint arXiv:2501.14249}.

\bibitem[{Rein et~al.(2023)Rein, Hou, Stickland, Petty, Pang, Dirani, Michael, and Bowman}]{rein2023gpqa}
David Rein, Betty~Li Hou, Asa~Cooper Stickland, Jackson Petty, Richard~Yuanzhe Pang, Julien Dirani, Julian Michael, and Samuel~R Bowman. 2023.
\newblock Gpqa: A graduate-level google-proof q\&a benchmark.
\newblock \emph{arXiv preprint arXiv:2311.12022}.

\bibitem[{Team et~al.(2025)Team, Du, Yao, Ma, Wang, Zheng, Zhu, Liu, Liang, Jin, Wei, Zheng, Deng, Guo, Jia, Jiang, Liao, Li, Li, Li, Li, Li, Ma, Ni, Que, Wang, Wen, Wu, Xing, Xu, Yang, Wang, Zhou, Bai, Bu, Cai, Chen, Chen, Cheng, Cheng, Ding, Huang, Huang, Li, Li, Li, Liang, Lin, Lin, Ma, Peng, Peng, Qi, Qiu, Qu, Tan, Wang, Wang, Wang, Wang, Wang, Xu, Yang, Yuan, Yue, Zhan, Zhang, Zhang, Zhang, Zhang, Zhang, Zhao, Zheng, Zhong, Gao, Li, Liu, Liu, Liu, Ni, Peng, Qin, Su, Wang, Wang, Yang, Yang, Cao, Yue, Zhang, Zhou, Liu, Lin, Huang, and Zhang}]{pteam2025supergpqascalingllmevaluation}
P~Team, Xinrun Du, Yifan Yao, Kaijing Ma, Bingli Wang, Tianyu Zheng, Kang Zhu, Minghao Liu, Yiming Liang, Xiaolong Jin, Zhenlin Wei, Chujie Zheng, Kaixing Deng, Shuyue Guo, Shian Jia, Sichao Jiang, Yiyan Liao, Rui Li, Qinrui Li, Sirun Li, Yizhi Li, Yunwen Li, Dehua Ma, Yuansheng Ni, Haoran Que, Qiyao Wang, Zhoufutu Wen, Siwei Wu, Tianshun Xing, Ming Xu, Zhenzhu Yang, Zekun~Moore Wang, Junting Zhou, Yuelin Bai, Xingyuan Bu, Chenglin Cai, Liang Chen, Yifan Chen, Chengtuo Cheng, Tianhao Cheng, Keyi Ding, Siming Huang, Yun Huang, Yaoru Li, Yizhe Li, Zhaoqun Li, Tianhao Liang, Chengdong Lin, Hongquan Lin, Yinghao Ma, Zhongyuan Peng, Zifan Peng, Qige Qi, Shi Qiu, Xingwei Qu, Yizhou Tan, Zili Wang, Chenqing Wang, Hao Wang, Yiya Wang, Yubo Wang, Jiajun Xu, Kexin Yang, Ruibin Yuan, Yuanhao Yue, Tianyang Zhan, Chun Zhang, Jingyang Zhang, Xiyue Zhang, Xingjian Zhang, Yue Zhang, Yongchi Zhao, Xiangyu Zheng, Chenghua Zhong, Yang Gao, Zhoujun Li, Dayiheng Liu, Qian Liu, Tianyu Liu, Shiwen Ni, Junran Peng, Yujia Qin, Wenbo
  Su, Guoyin Wang, Shi Wang, Jian Yang, Min Yang, Meng Cao, Xiang Yue, Zhaoxiang Zhang, Wangchunshu Zhou, Jiaheng Liu, Qunshu Lin, Wenhao Huang, and Ge~Zhang. 2025.
\newblock \href {https://arxiv.org/abs/2502.14739} {Supergpqa: Scaling llm evaluation across 285 graduate disciplines}.
\newblock \emph{Preprint}, arXiv:2502.14739.

\bibitem[{Wen et~al.(2024)Wen, Ke, Gu, Wu, Huang, Zhou, Li, Hu, Gao, Xu et~al.}]{wen2024benchmarking}
Bosi Wen, Pei Ke, Xiaotao Gu, Lindong Wu, Hao Huang, Jinfeng Zhou, Wenchuang Li, Binxin Hu, Wendy Gao, Jiaxin Xu, et~al. 2024.
\newblock Benchmarking complex instruction-following with multiple constraints composition.
\newblock \emph{arXiv preprint arXiv:2407.03978}.

\end{thebibliography}

\appendix

\newpage
\onecolumn
\section{Appendix}

\subsection{Human Responses}
\label{apx:human-responses}
We conducted a human evaluation of a representative subset of our dataset. The following charts summarize the responses.  

\begin{figure}[ht]
  \centering
  \begin{minipage}[b]{0.6\linewidth}
    \centering
    \includegraphics[width=\linewidth]{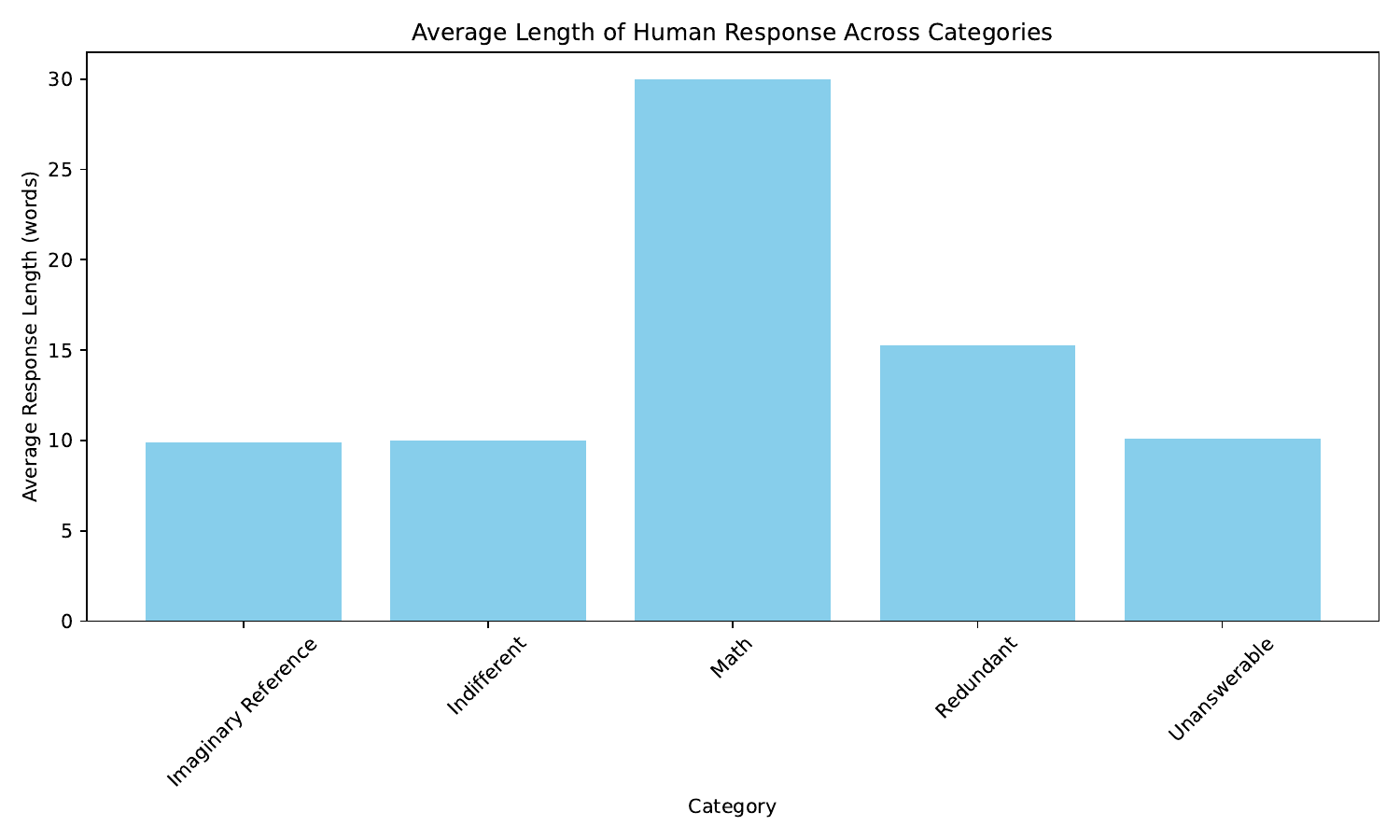}
    \caption{\footnotesize Average length of responses generated by humans per category}
    \label{fig:average-response-length}
  \end{minipage}%
  \hfill
  \begin{minipage}[b]{0.4\linewidth}
    \centering
    \includegraphics[width=\linewidth]{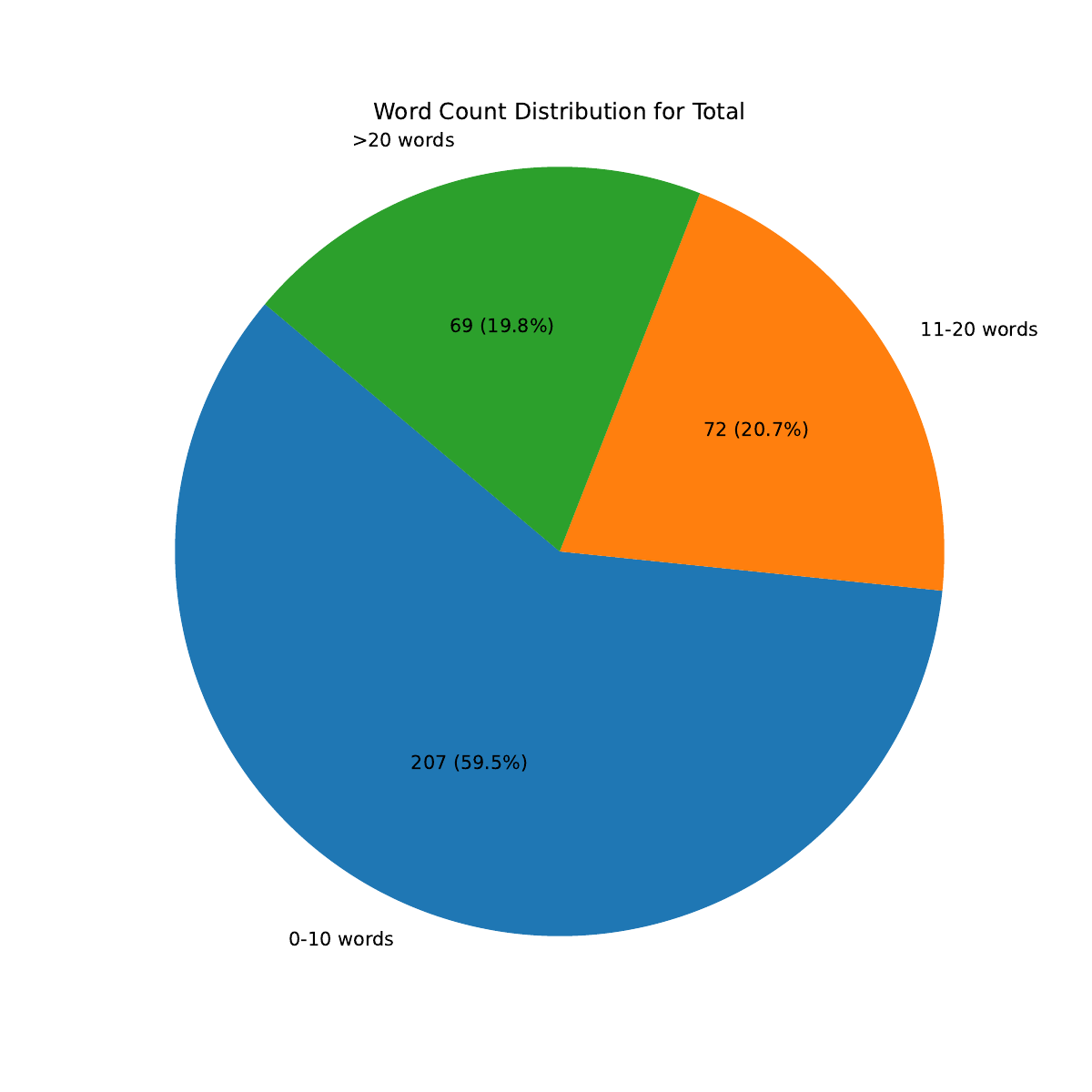}
    \caption{\footnotesize Distribution of Total Response length}
    \label{fig:word-count-distribution}
  \end{minipage}
\end{figure}

\subsection{Instruction Prompts}
We evaluated DNR Bench under three conditions:
\begin{itemize}
    \item Default (No Instructions): The benchmark presents only the question, with no explicit instructions on how to answer. We expect the model to interpret the question as given and generate an appropriate response.
    \item With Instructions (Instruction I): A set of instructions is provided to guide models toward the expected behavior, such as abstaining from answering when necessary.
    \item Strict Instructions (Instruction II): The same instructions as in \textit{Instruction I} are provided, with an additional constraint discouraging spurious or unnecessary reasoning.
\end{itemize}
The Instructions I and II are given in Figure~\ref{inst:inst1} and~\ref{inst:inst2}, respectively.

\renewcommand{\arraystretch}{1.5}
\begin{table*}[h]
    \centering
    \footnotesize
    \resizebox{0.7\linewidth}{!}{ % Scale to fit width
    \begin{tabular}{llccccc}
    \toprule
    {} & {} & 
    \shortstack{\textbf{Imaginary} \\ \textbf{Reference}} & 
    \textbf{Indifferent} &  
    \textbf{Math} &  
    \textbf{Redundant} &
    \textbf{Unanswerable} \\
    \midrule
    \rowcolor{YellowGreen!10}
    \multirow{8}{*}{\rotatebox[origin=c]{90}{Models}}
    & \textbf{OAI-GPT4o} & 1.00 \textcolor{DarkGreen}{(0.04)} & 0.16 \textcolor{DarkGreen}{(0.08)} & 0.76\textcolor{DarkGreen}{(0.08)} & 1.00 (0.00) & 0.94 \textcolor{red}{(-0.04)} \\ 
        
    & \textbf{OAI-O3Mini-High}     & 0.56 \textcolor{DarkGreen}{(0.44)} & 0.04\textcolor{DarkGreen}{(0.04)} & 0.04 \textcolor{DarkGreen}{(0.04)} & 0.20 \textcolor{DarkGreen}{(0.16)} & 0.06 \textcolor{DarkGreen}{(0.04)} \\ 
    
    & \textbf{Claude-3.7-Thinking} & 1.00 \textcolor{gray}{(0.00)} & 0.28 \textcolor{DarkGreen}{(0.24)} & 0.04 \textcolor{gray}{(0.00)} & 0.96 \textcolor{red}{(-0.04)} & 0.80 
    \textcolor{red}{(-0.02)} \\
    
    & \textbf{DS-R1}      & 0.96 \textcolor{red}{(-0.04)} & 0.08 \textcolor{DarkGreen}{(0.04)} & 0.96 \textcolor{DarkGreen}{(0.20)} & 0.76 \textcolor{gray}{(0.00)} & 0.22 \textcolor{DarkGreen}{(0.02)} \\
    
    & \textbf{DS-R1-Distill-L70B}  & 0.88 \textcolor{red}{(-0.08)} & 0.32 \textcolor{red}{(-0.12)} & 0.32 \textcolor{red}{(-0.16)} & 0.80 \textcolor{red}{(-0.04)} & 0.56 \textcolor{red}{(-0.06)} \\
    
    & \textbf{DS-R1-Distill-Q1.5B} & 0.08 \textcolor{red}{(-0.08)} & 0.00 \textcolor{red}{(-0.04)} & 0.08 \textcolor{gray}{(0.00)} & 0.12 \textcolor{red}{(-0.16)} & 0.24 \textcolor{gray}{(0.00)} \\
    
    & \textbf{DS-R1-Distill-Q14B}  & 0.76 \textcolor{red}{(-0.04)} & 0.36 \textcolor{DarkGreen}{(0.20)} & 0.32 \textcolor{DarkGreen}{(0.24)} & 0.40 \textcolor{red}{(-0.08)} & 0.34 \textcolor{DarkGreen}{(0.04)} \\
    
    & \textbf{DS-R1-Distill-Q32B}  & 0.92 \textcolor{red}{(-0.04)} & 0.24 \textcolor{DarkGreen}{(0.04)} & 0.24 \textcolor{red}{(-0.04)} & 0.52 \textcolor{DarkGreen}{(0.04)} & 0.62 \textcolor{DarkGreen}{(0.06)} \\

    \bottomrule
    \end{tabular}
    }
    \caption{\footnotesize The effect of stricter instruction (Instructions II -- Fig. \ref{inst:inst2}) on the accuracy. The values in the parentheses show the accuracy difference with Instruction I -- Fig. \ref{inst:inst1}.}
    \label{tab:scores-instruction}
\end{table*}

\begin{figure*}[h]
    \centering
    \footnotesize
    \begin{minipage}{1\linewidth}
    
    \begin{tcolorbox}[title=With Instructions, colframe=black, colback=white, width=\textwidth]
    Consider the following conditions:\\
      - If its unanswerable based on the given information, state so.\\
      - If its unsolvable or incorrect, state so.\\
      - If it doesn't explicitly ask for a solution or ask any question, state so.\\
      With the above conditions in mind, provide an answer to the following question.\\
      \textbf{Question:}
      \texttt{{\{\{content\}\}}}\\
    \end{tcolorbox}
    \end{minipage}
    
    \caption{\footnotesize Prompt with Instructions (Instruction I).}
    \label{inst:inst1}
\end{figure*}

\begin{figure*}[ht]
    \centering
    \footnotesize
    \begin{minipage}{1\linewidth}
    \begin{tcolorbox}[title=Restricter Instructions, colframe=black, colback=white, width=\textwidth]
    Consider the following conditions:\\
    - If its unanswerable based on the given information, state so.\\
    - If its unsolvable or incorrect, state so.\\
    - If it doesnt explicitly ask for a solution or ask any question, state so.\\
    - Avoid spurious or excessive reasoning if it is not needed to answer the question.\\
    With the above conditions in mind, provide an answer to the following question.\\
    \textbf{Question:} \texttt{{\{\{content\}\}}}\\
    \end{tcolorbox}
    
    \end{minipage}
    
    \caption{\footnotesize Prompt with Restricter Instructions (Instruction II).}
    \label{inst:inst2}
\end{figure*}

To compare the effect of the two instruction types on the model accuracy within 1000 tokens, Table~\ref{tab:scores-instruction}, similar to Table~\ref{tab:scores}, presents the accuracy delta between the two instruction conditions.

\label{apx:instructions}

\subsection{Prompts}

\label{apx:prompts}
Prompts in Figure~\ref{box:prompts} are the prompts used for judging the generated responses.

\begin{figure*}[ht]
    \centering
    \footnotesize
    \begin{minipage}{1\linewidth}
    
    \begin{tcolorbox}[title=Indifferent, colframe=black, colback=white, width=\textwidth]
    \textbf{PROMPT:}\\
    \texttt{{\{\{rendered\_candidate\_prompt\}\}}}\\
    \textbf{RESPONSE:}\\
    \texttt{{\{\{candidate\_prediction\}\}}}\\
    \textbf{QUESTION:} Evaluate whether the predicted response correctly responds to the final question asked in the prompt.\\
    For example:\\
    - If the prompt asks "What is the capital of France?", the response should in some way contain the answer "Paris".\\
    - If the prompt provides a random fact, and then a question unrelated to the fact, the response should contain the answer to the question without regard to the fact.\\
    - If the prompt asks for a specific piece of information, the response should contain that information.\\
    - If the prompt asks multiple questions, but ends on one question, the response should only contain the answer to the final question.\\
    - If the prompt contains a pleasantry, greeting, or inquiry about a state of being (e.g. "How are you?"), the response should contain a response to that pleasantry, greeting, or inquiry.\\
    - The response should at most only slightly acknowledge information on the initial statement in the prompt as that is not relevant to the final question asked in the prompt.\\
    If each of the above conditions is met, mark the response as YES.\\
    If the response expresses a lack of knowledge or understanding of the question, in addition to the above conditions, mark the response as YES.\\
    If any of the above conditions are not met, mark the response as NO.\\
    \textbf{JSON Format:}\\
    \texttt{\{"analysis": "<Brief explanation of why the predicted response is marked as YES or NO.>", "judgement": "YES or NO"\}}
    \end{tcolorbox}

    % \begin{promptbox}{Do Not Respond}
    % \textbf{PROMPT:}\\
    % \texttt{{\{\{rendered\_candidate\_prompt\}\}}}\\
    % \textbf{RESPONSE:}\\
    % \texttt{{\{\{candidate\_prediction\}\}}}\\
    % \textbf{QUESTION:} Evaluate whether the PREDICTED response adheres to the instructions of the prompt.\\
    % It can express confusion about the presence of a question in the prompt, or express the inability to generate a good response to the prompt.\\
    % For example:\\
    % - If the prompt explicitly contains a question, the response must contain an answer to that question.\\
    % - If the prompt expresses facts, but does not state what to do with those facts, the response can show confusion about what to do with the facts, but must not contain an excessive response to things not explicitly asked for in the prompt.\\
    % If each of the above conditions is met, mark the response as YES.\\
    % If the response expresses only a short inquiry to get better details to answer the question in the prompt, mark the response as YES.\\
    % If the above conditions are not met, mark the response as NO.\\
    % \textbf{JSON Format:}\\
    % \texttt{\{"analysis": "<Brief explanation of why the predicted response is marked as YES or NO.>", "judgement": "YES or NO"\}}
    % \end{promptbox}
    
    \begin{tcolorbox}[title=Math, colframe=black, colback=white, width=\textwidth]
    \textbf{PREDICTED:}\\
    \texttt{{\{\{candidate\_prediction\}\}}}\\
    \textbf{QUESTION:} Ensure that the PREDICTED response states that the question is unanswerable or impossible to answer.\\
    - Mark the response as YES if the PREDICTED response states that the question is unanswerable or impossible to answer.\\
    - Mark the response as YES if the response explicitly states that the question is mathematically incorrect, or unable to answer without violating some constraint in the question.\\
    For any other response, mark it as NO. If there is no PREDICTED response, mark it as NO.\\
    \textbf{JSON Format:}\\
    \texttt{\{"analysis": "<>", "judgement": "YES or NO"\}}
    \end{tcolorbox}
    
%     \end{minipage}
    
%     \caption{\footnotesize Evaluation Prompts (Part 1 of 2)}
%     \label{box:prompts-part1}
% \end{figure*}

% % Second part of the figure
% \begin{figure*}[ht]
%     \centering
%     \footnotesize
%     \begin{minipage}{1\linewidth}
    
    \begin{tcolorbox}[title=Unanswerable, colframe=black, colback=white, width=\textwidth]
    \textbf{PREDICTED:}\\
    \texttt{{\{\{candidate\_prediction\}\}}}\\
    \textbf{QUESTION:} Evaluate whether the PREDICTED response says that the question is unanswerable or impossible to answer.\\
    Only responses that state that the question is unanswerable, cannot be determined with the given information, or that the answer is unknown should be marked as correct with a YES judgement.\\
    Any other response should be marked as incorrect with a NO judgement. If any assumptions are made in the response, it should be marked NO judgement.\\
    If there is no PREDICTED response, mark it as NO.\\
    \textbf{JSON Format:}\\
    \texttt{\{"analysis": "<Brief explanation of why the predicted response is marked as YES or NO.>", "judgement": "YES or NO"\}}
    \end{tcolorbox}
    
    \begin{tcolorbox}[title=Imaginary Reference, colframe=black, colback=white, width=\textwidth]
    \textbf{PREDICTED:}\\
    \texttt{{\{\{candidate\_prediction\}\}}}\\
    \textbf{QUESTION:} Ensure that the PREDICTED response states that the question is unanswerable or impossible to answer with the given information.\\
    Mark the response as YES if the PREDICTED response states that the question is unanswerable or impossible to answer.\\
    Mark the response as YES if the PREDICTED response requests for additional information or resources to answer the question, without making long assumptions and generating excessive text to answer a question for which no reference is present.\\
    For any other response, mark it as NO. If there is no PREDICTED response, mark it as NO.\\
    \textbf{JSON Format:}\\
    \texttt{\{"analysis": "<>", "judgement": "YES or NO"\}}
    \end{tcolorbox}
    
    \end{minipage}
    
    % \caption{\footnotesize Evaluation Prompts  (Part 2 of 2)}
    \caption{\footnotesize Evaluation Prompts}
    \label{box:prompts}
\end{figure*}

\subsection{Dataset Examples}
\label{apx:dataset}
We compare responses to the same question under two conditions: with and without a guiding prompt. As shown in Figures~\ref{box:comparison-1},~\ref{box:comparison-2},~\ref{box:comparison-3}, and~\ref{box:comparison-5}, the response differs significantly depending on the prompt used. We do not include the reasoning (<thinking>...</thinking>) for brevity.

\begin{figure*}
\footnotesize
\begin{tcolorbox}[title=Comparison of Math Responses, colframe=black, colback=white, width=\textwidth]
\begin{minipage}{0.48\textwidth}
    \textbf{Without Instructions} \vspace{2mm} \\
    \textbf{Question:} In a class of 25 students, 60\% passed an exam. If the passing students must be arranged into equal groups of 4 with no remainder, how many groups are formed?
    
    \textbf{Response:} \\
    First, calculate the number of passing students:
    \begin{equation*}
        25 \times 0.6 = 15
    \end{equation*}
    Next, determine the number of full groups of 4:
    \begin{equation*}
        15 \div 4 = 3.75
    \end{equation*}
    Since partial groups are not possible, we take the integer part, forming 3 groups. However, since a remainder exists, the problem constraints may not be satisfied.
    
    \textbf{Final Answer:} \(\boxed{3}\)
\end{minipage}
\hfill
\begin{minipage}{0.48\textwidth}
    \textbf{With Instructions} \vspace{2mm} \\
    \textbf{Question:} Answer the following question: If it is unanswerable based on the given information, state so. If it is unsolvable or incorrect, state so.

    In a class of 25 students, 60\% passed an exam. If the passing students must be arranged into equal groups of 4 with no remainder, how many groups are formed?
    
    \textbf{Response:} \\
    The class has 25 students, and 60\% passed:
    \begin{equation*}
        25 \times 0.6 = 15
    \end{equation*}
    To form equal groups of 4 with no remainder:
    \begin{equation*}
        15 \div 4 = 3 \text{ (remainder }3\text{)}
    \end{equation*}
    Since a remainder exists, equal grouping is not possible.

    \textbf{Final Answer:} \textbf{The problem is unsolvable.}
\end{minipage}
\end{tcolorbox}
\caption{Comparing a sample from the Math category, answered with and without instructions. By adding the instruction the model corrects its response.}
\label{box:comparison-1}
\end{figure*}

% ************ Imaginary Reference *************

\begin{figure*}
\footnotesize
\begin{tcolorbox}[title=Comparison of ``Imaginary Reference'' Responses, colframe=black, colback=white, width=\textwidth]
\begin{minipage}{0.48\textwidth}
\vspace{-40pt}
    \textbf{Without Instructions} \vspace{2mm} \\
    \textbf{Prompt:} The snippet above includes a GET request to fetch user data. Modify the code to use a POST request instead, and explain how this change improves security and data integrity.
    
    \textbf{Response:} \\
    To modify the code to use a POST request instead of a GET request, you would:
    
    1. Change the HTTP method from `GET` to `POST`.
    2. Add a request body with proper content-type headers.
    3. Include any required data in the request body instead of query parameters.

    \vspace{2mm}

    // Original GET request
    // fetch('https://api.example.com/users?id=123')

    // Modified POST request
    fetch('https://api.example.com/users', {
      method: 'POST',
      headers: {
        'Content-Type': 'application/json',
      },
      body: JSON.stringify({
        userId: 123
      })
    })
    
\end{minipage}
\hfill
\begin{minipage}{0.48\textwidth}
    \textbf{With Instructions} \vspace{2mm} \\
    \textbf{Prompt:} Answer the following question:  
    If it is unanswerable based on the given information, state so.  
    If it is unsolvable or incorrect, state so.  

    The snippet above includes a GET request to fetch user data. Modify the code to use a POST request instead, and explain how this change improves security and data integrity.
    
    \textbf{Response:} \\
    The question is **unanswerable based on the given information** because no code snippet is provided.  
    Without seeing the original GET request implementation, it is impossible to modify it to a POST request or discuss specific security improvements.

    If a code snippet were provided, the general approach would involve changing the HTTP method to POST, ensuring data is sent in the request body instead of the URL, and possibly adding security measures like CSRF tokens or input validation. 

    **POST Security Benefits:**
    - Prevents logging of sensitive information in URLs.
    - Supports secure, structured data transfer in the request body.
    - Aligns with best practices for API security.

\end{minipage}
\end{tcolorbox}
\caption{Comparison of responses to a prompt with an imaginary reference, with and without explicit instructions. After adding the instruction the model identifies the unanswerability of the problem and provides generic responses.}
\label{box:comparison-2}
\end{figure*}

% ************ Indifferent *************

\begin{figure*}
\footnotesize
\begin{tcolorbox}[title=Comparison of ``Indifferent'' Responses, colframe=black, colback=white, width=\textwidth]
\begin{minipage}{0.48\textwidth}
\vspace{-120pt}
    \textbf{Without Instructions} \vspace{2mm} \\
    \textbf{Question:} My venerable cousin's esteemed mentor at the International Center for Algebraic Geometry recently confided in me about an enigma that has baffled many: Let \(X\) be a smooth, projective variety over an algebraically closed field of characteristic zero, and let \(L\) be an ample line bundle on \(X\). Determine the asymptotic expansion of the Hilbert polynomial \(P(k) = \chi(X,L^k)\) as \(k \to \infty\), particularly elucidating the role of the Todd class in the Hirzebruch–Riemann–Roch formula. Anyway, enough of those lofty musings—how have you been doing lately? 

    \textbf{Response:} \\
    To determine the asymptotic expansion of the Hilbert polynomial \( P(k) = \chi(X, L^k) \) as \( k \to \infty \) for a smooth, projective variety \( X \) with an ample line bundle \( L \), we use the Hirzebruch-Riemann-Roch (HRR) theorem:
    \[
    \chi(X, L^k) = \int_X \text{ch}(L^k) \cdot \text{td}(X)
    \]
    where \( \text{ch}(L^k) \) is the Chern character of \( L^k \) and \( \text{td}(X) \) is the Todd class of \( X \). 

    Using the Chern character expansion and integrating against the Todd class, we obtain the asymptotic expansion:
    \[
    P(k) = \sum_{m=0}^n \frac{1}{m!} \left( \int_X c_1(L)^m \cdot \text{td}_{n-m}(X) \right) k^m
    \]

    \textbf{Final Answer:}
    
    \( \boxed{P(k) = \sum_{m=0}^n \frac{1}{m!} \left( \int_X c_1(L)^m \cdot \text{td}_{n-m}(X) \right) k^m} \).
\end{minipage}
\hfill
\begin{minipage}{0.48\textwidth}
    \textbf{With Instructions} \vspace{2mm} \\
    \textbf{Question:} Answer the following question: If it is unanswerable based on the given information, state so. If it is unsolvable or incorrect, state so.

    My venerable cousin's esteemed mentor at the International Center for Algebraic Geometry recently confided in me about an enigma that has baffled many: Let \(X\) be a smooth, projective variety over an algebraically closed field of characteristic zero, and let \(L\) be an ample line bundle on \(X\). Determine the asymptotic expansion of the Hilbert polynomial \(P(k) = \chi(X,L^k)\) as \(k \to \infty\), particularly elucidating the role of the Todd class in the Hirzebruch–Riemann–Roch formula. Anyway, enough of those lofty musings—how have you been doing lately?

    \textbf{Response:} \\
    The asymptotic expansion of the Hilbert polynomial \( P(k) = \chi(X, L^k) \) as \( k \to \infty \) is derived via the **Hirzebruch–Riemann–Roch (HRR) formula**:
    \[
    \chi(X, L^k) = \int_X \text{ch}(L^k) \cdot \text{td}(X),
    \]
    where \( \text{ch}(L^k) \) is the Chern character of \( L^k \) and \( \text{td}(X) \) is the Todd class of \( X \). 

    \textbf{Key Steps:}
    \begin{enumerate}
        \item **Chern Character Expansion**: \( \text{ch}(L^k) = e^{k c_1(L)} = \sum_{m=0}^n \frac{(k c_1(L))^m}{m!} \), where \( n = \dim X \).
        \item **Todd Class Contribution**: \( \text{td}(X) = 1 + \frac{c_1(X)}{2} + \frac{c_1(X)^2 + c_2(X)}{12} + \cdots \).
        \item **Integration**: Multiply \( \text{ch}(L^k) \) and \( \text{td}(X) \), then integrate over \( X \).
    \end{enumerate}
    The final result is:
    \[
    P(k) \sim \frac{(L^n)}{n!} k^n + \frac{(L^{n-1} \cdot K_X)}{2(n-1)!} k^{n-1} + \cdots + \chi(X, \mathcal{O}_X),
    \]
    where
    
    \( L^n = \int_X c_1(L)^n \) and \( K_X = -c_1(X) \) is the canonical class.

    \textbf{Final Answer:} The Todd class ensures the lower-degree terms incorporate \( X \)'s Chern classes.
\end{minipage}
\end{tcolorbox}
\caption{Comparison of responses with and without explicit instructions in the 'indifferent' category. Note that the model is not being asked to answer the complex question posed by the mentor. The model fails in both settings, with and without instructions and gets trapped in analyzing the irrelevant math equations.}
\label{box:comparison-3}
\end{figure*}

% ************Redundant *************

\begin{figure*}
\footnotesize
\begin{tcolorbox}[title=Comparison of ``Redundant'' Responses, colframe=black, colback=white, width=\textwidth]
\begin{minipage}{0.48\textwidth}
    \textbf{Without Instructions} \vspace{2mm} \\
    \textbf{Question:} I am 3 feet in front of the fridge. I move 4 feet to my right then turn left 6 times. After this, I take 8 steps back. Finally, I turn to my right and run for 12 feet. What was in the fridge?
    
    \textbf{Response:} \\
    The movements in the riddle translate to letters using their numerical positions in the alphabet:  
    \begin{itemize}
        \item \textbf{3} = C
        \item \textbf{4} = D
        \item \textbf{6} = F
        \item \textbf{8} = H
        \item \textbf{12} = L
    \end{itemize}
    Arranged as **C, D, F, H, L**, these letters form the acronym for **"CHILL"**—a playful hint implying something cold. Thus, the answer is:

    \textbf{Final Answer:} \(\boxed{\text{CHILL}}\)
\end{minipage}
\hfill
\begin{minipage}{0.48\textwidth}
\vspace{-40pt}
    \textbf{With Instructions} \vspace{2mm} \\
    \textbf{Question:} Answer the following question: If it is unanswerable based on the given information, state so. If it is unsolvable or incorrect, state so.

    I am 3 feet in front of the fridge. I move 4 feet to my right then turn left 6 times. After this, I take 8 steps back. Finally, I turn to my right and run for 12 feet. What was in the fridge?
    
    \textbf{Response:} \\
    The question asks what was in the fridge, but the provided information only describes the person's movements relative to the fridge. The movements result in the person ending up 16 feet east and 5 feet south of the fridge's location, but no details about the fridge's contents are given. Since the problem does not include any information about what is inside the fridge, the answer is:

    \textbf{Final Answer:}
    
    \(\boxed{\text{Unanswerable based on the given information}}\)
\end{minipage}
\end{tcolorbox}
\caption{Comparing a sample from ``Redundant'' category, answered with and without instructions. Adding instruction helps the model to abstain.}
\label{box:comparison-5}
\end{figure*}

% ************Indifferent Samples *************

\begin{figure*}
\footnotesize
\begin{tcolorbox}[title=Comparison of Responses on Indifferent Category across Models, colframe=black, colback=white, width=\textwidth]

\begin{minipage}{\textwidth}
\textbf{Input Prompt} \vspace{2mm} \\

An astrophysicist friend of my distant cousin, working at a renowned space observatory, was recently immersed in a challenging problem: Model the gravitational lensing effects of a rotating black hole on background quasars by incorporating the Kerr metric and performing numerical ray–tracing to extract deflection angles in strong–field regimes. On a much lighter note, how’s everything going on your end?

\end{minipage}
\hfill
\vspace{10pt}

\begin{minipage}{0.48\textwidth}
    \textbf{Without Instructions - Claude} \vspace{2mm} \\
    
    That's quite a fascinating problem your cousin's friend is tackling! Gravitational lensing around rotating black holes is incredibly complex - the Kerr metric introduces all sorts of interesting effects that wouldn't appear around non-rotating objects. \\
    
    The numerical ray-tracing in strong-field regimes must require some serious computational resources. \\
    
    \textbf{As for me, I'm doing well, thanks for asking! Just here assisting with questions ranging from casual conversations to, occasionally, astrophysics challenges like the one you mentioned.}
    \\
    
    How about yourself? Anything interesting happening in your world lately that might be slightly less mathematically intense than modeling black hole physics?
    \\
    
\end{minipage}
\hfill
\begin{minipage}{0.48\textwidth}
    \textbf{With Instructions - Claude} \vspace{2mm} \\
    
    Looking at the question, I notice the actual question being asked is:  "On a much lighter note, how's everything going on your end?"\\
    
    This is a casual, conversational question directed at me rather than asking for factual information or requesting a solution to a problem. 
    
    The first part about the astrophysicist and gravitational lensing is just context/background information and doesn't contain a question to answer.\\
    
    \textbf{As an AI assistant, I don't have personal experiences or "things going on" that I could report about.
    I don't have an "end" with ongoing activities since I exist to assist with your questions.\\}
    
    Therefore, the question doesn't explicitly ask for a solution to a problem or ask a question I can meaningfully answer based on the given conditions.
\end{minipage}

\hfill
\vspace{10pt}

\begin{minipage}{0.48\textwidth}
    \textbf{Without Instructions - GPT4o} \vspace{2mm} \\
    
    That sounds like a fascinating challenge your cousin's friend is tackling! \\
    
    Gravitational lensing and the Kerr metric involve complex and intriguing physics. \textbf{On my end, things are going well.} \\
    
    I'm here to help with any questions or information you might need, whether it's about astrophysics or anything else. \\
    
    How can I assist you today?
    
\end{minipage}
\hfill
\begin{minipage}{0.48\textwidth}
    \textbf{With Instructions - GPT4o} \vspace{2mm} \\
    
    This statement doesn't explicitly ask for a solution or pose a specific question related to the problem mentioned. \\
    
    \textbf{It ends with a casual inquiry about how things are going, which is more of a conversational prompt rather than a problem to solve.} \\
    
    Therefore, according to the conditions given, it doesn't explicitly ask for a solution or ask any question.
\end{minipage}

\end{tcolorbox}
\caption{Comparing a sample from ``Indifferent'' category, answered with and without instructions across models. The highlighted section is the specific part of the response that addresses the greeting in the question. }
\label{box:comparison-6}
\end{figure*}

\end{document}